\documentclass[10pt,twocolumn,letterpaper]{article}

\usepackage{iccv}
\usepackage{times}
\usepackage{epsfig}
\usepackage{graphicx}
\usepackage{amsmath}
\usepackage{amssymb}
\usepackage{booktabs}
\usepackage{multirow}
\usepackage{makecell}
\usepackage{bbding}
\usepackage[pagebackref=true,breaklinks=true,letterpaper=true,colorlinks,bookmarks=false]{hyperref}

\iccvfinalcopy % *** Uncomment this line for the final submission

\def\iccvPaperID{9662} % *** Enter the ICCV Paper ID here

\ificcvfinal\fi

\begin{document}

%%%%%%%%% TITLE
\title{MGTAB: A Multi-Relational Graph-Based Twitter Account Detection Benchmark}

\author{Shuhao Shi$^{1,\dag}$, Kai Qiao$^{1,\dag}$, Jian Chen$^{1}$ Shuai Yang$^{1}$, Jie Yang$^{1}$,\\
Baojie Song$^{1}$, Linyuan Wang$^{1}$, Bin Yan$^{1,*}$\\
$^1$Henan Key Laboratory of Imaging and Intelligence Processing,\\ PLA strategy support force information engineering university\\
\vspace*{-0.3cm}
}

\maketitle
\ificcvfinal\thispagestyle{empty}\fi

%%%%%%%%% ABSTRACT
\begin{abstract}
\vspace*{-0.2cm}
The development of social media user stance detection and bot detection methods rely heavily on large-scale and high-quality benchmarks. However, due to low annotation quality and incomplete user relationships, existing benchmarks often impede graph-based account detection research. To counter these issues, we introduce the Multi-Relational Graph-Based Twitter Account Detection Benchmark (MGTAB), which is the first graph-based benchmark to jointly annotate bot and stance. To the best of our knowledge, MGTAB was built based on the largest original data in the field, comprising over 1.55 million users and 130 million tweets. MGTAB includes 10,199 expert-annotated users and 7 types of relationships, ensuring both high-quality annotation and diversified relations. We also extracted the 20 user property features with the greatest information gain and user tweet features as the user features. Moreover, we conducted a comprehensive evaluation of MGTAB and other public datasets. Our experiments revealed that graph-based approaches are generally more efficacious than feature-based approaches and perform better when incorporating multiple relations. By analyzing experiment results, we identify effective approaches for account detection and suggest potential future research directions in this field. Our benchmark and standardized evaluation procedures are freely available at: \small{\url{https://github.com/GraphDetec/MGTAB}}.
\vspace*{-0.35cm}
\end{abstract}

%%%%%%%%% BODY TEXT
\section{Introduction}

\label{sec:intro}
With the continuous development of the Internet, social networks have become an essential part of people's daily social life. Twitter is one of the most widely-used social networks globally, offering online news and information exchange to billions of users worldwide. Due to the availability, many account detection benchmarks are constructed based on Twitter data~\cite{Alpher10,Alpher11,Alpher12,Alpher21}.

Stance detection seeks to identify the user's stance on a topic or claim, and is a key technique in applications such as fake news detection~\cite{Alpher25,Alpher27}, claims validation~\cite{Alpher09,Alpher30}, and assessing public opinion on social media. Bot detection is indispensable for recognizing manipulation of information on social media. Social bots are automated user accounts operated by computer programs~\cite{Alpher35} and are often used to abuse social media platforms~\cite{Alpher04,Alpher05} to manipulate public opinion~\cite{Alpher04,Alpher07,Alpher10,Alpher35}.

Most account detection methods typically utilize only a fraction of the available information in social media, such as posts and registration information, for classification. Rarely consider the connection between users~\cite{Alpher08}, making it difficult to guarantee accurate detection. In stance detection, silent users rarely post directly but instead express their stance through behaviors, such as following others and favoring posts~\cite{Alpher08}. Unfortunately, most studies solely focus on the posting content of active users, neglecting silent users~\cite{Alpher08}. The features of social graphs must be employed to more effectively detect the silent users' stance~\cite{Alpher09}. Regarding bot detection, since most studies ignore bots' social graph features, bots can utilize intricate strategies to simulate genuine users and evade feature-based detection methods~\cite{Alpher04}.

Recent work~\cite{Alpher06,Alpher24,Alpher44} in account detection has focused on exploiting relationships between users, with performance improvement compared to feature-based approaches. However, the existing datasets have several drawbacks to supporting graph-based methods, as follows:

\begin{itemize}
\vspace{-0.2cm}
\item \textbf{(a) Low annotation quality.} Previous account detection datasets were mostly annotated by crowdsourcing, while crowdworkers' lack of domain knowledge resulted in significant noise in the annotation~\cite{Alpher12}.
\vspace{-0.2cm}
\item \textbf{(b) Incomplete user relationships.} None of the stance detection datasets explicitly provide the graph structure among users, and only the bot detection datasets Cresci-15~\cite{Alpher10}, TwiBot-20~\cite{Alpher11}, and TwiBot-22~\cite{Alpher12} contain explicit graph structures. Moreover, Cresci-15 and TwiBot-20 contain only 2 types of user relationships, which is insufficient for graph-based detection methods.
\vspace{-0.2cm}
\item \textbf{(c) Complex user information.} Social media user information is diverse and voluminous, but most information has little effect on account detection. Existing datasets lack the extraction and organization of essential user information, making account detection a difficult problem.
\end{itemize}

To address the shortcomings above, we presented Multi-Relational Graph-Based Twitter Account Detection Benchmark (MGTAB), a large expert-annotated dataset for stance and bot detection. MGTAB contains 10,199 users manually annotated by experts and 400,000 closely related unannotated users. Further, MGTAB extracted 20 most effective user property features by calculating the information gain (IG) and user tweet features. Finally, MGTAB simplified the social graph and constructed a user network with 7 types of relationships. The contributions of this paper are as follows:
\begin{itemize}
\item We presented MGTAB, a large-scale expert-annotated benchmark for stance detection and bot detection. All annotations are carried out by experts and improve annotation quality by cross-validation. The annotation quality has been substantially improved compared to the previous dataset.
\vspace{-0.2cm}
\item We released the first normalized dataset containing the property features, user tweet features and 7 types of user relationships. We constructed a user-level social graph that can be applied to state-of-the-art graph-based account detection methods, making account detection simpler. The release of the MGTAB dataset will facilitate the development of new methods for graph-based account detection.
\vspace{-0.2cm}
\item To build MGTAB, we collect over 1.55 million Twitter users and 135 million tweets. To the best of our knowledge, it is the biggest data in the domain. We carried out meticulous data cleaning, retaining 400,000 closely related unlabeled users, which supports semi-supervised learning to merge with account detection research.
\vspace{-0.2cm}
\item Our experiments show that graph-based detection methods are more effective than feature-based methods in most cases. In addition, We found that the performance of graph-based approaches improved when multiple relationships were introduced. The results suggest that future research should prioritize the utilization of multiple relationships.
\end{itemize}

\section{Related Work}
\label{sec:rela}
\subsection{Stance Detection}
The existing stance detection methods can be divided into feature-based methods and graph-based methods.

\noindent
\textbf{Feature-based methods.}
Previous research works ~\cite{Alpher13,Alpher14,Alpher15} used machine learning algorithms and deep learning methods such as Support Vector Machines (SVM), Recurrent Neural Networks (RNNs)~\cite{Alpher13}, and Convolutional Neural Networks (CNNs) to automatically learn latent features from a large amount of raw data. Several recent works ~\cite{Alpher16,Alpher17,Alpher18,Alpher27,Alpher28} focused on the use of bidirectional encoder representations from transformers (BERT)~\cite{Alpher02} on stance detection. Ghosh \etal~\cite{Alpher19} explored stance detection based on transfer learning, and Li \etal~\cite{Alpher16} explored BERT-based data augmentation models.

\noindent
\textbf{Graph-based methods.}
Most studies on stance detection have focused on text-based features~\cite{Alpher13,Alpher21,Alpher28}. However, recent work has demonstrated the effectiveness of using user network graphs as features~\cite{Alpher09,Alpher23}. Graph Neural Networks (GNNs)~\cite{Alpher20,Alpher42} have become the preferred model for account detection due to their ability to process graph information. Li \etal~\cite{Alpher24} first achieved stance and rumor detection using a GNN-based architecture that efficiently captured user interaction characteristics. However, the lack of graph structure in existing stance detection datasets hinders the development of graph-based detection methods.

\noindent
\textbf{Stance Detection Dataset.}
Tab.~\ref{tab:1} summarizes the existing Twitter stance detection datasets. SemEval-2016 T6 dataset~\cite{Alpher21} is the first dataset for Twitter stance detection, which contains topic-tweet pairs annotated by crowdworkers. SemEval-2019 T7~\cite{Alpher25} contains rumors about various incidents from Reddit posts and tweets. COVID-19-Stance~\cite{Alpher31} consists of manually annotated tweets covering users’ stances towards four targets relevant to COVID-19 health mandates. COVIDLies~\cite{Alpher26}, COVMis-Stance~\cite{Alpher27} are also COVID-related datasets. P-STANCE~\cite{Alpher28} is a large stance detection dataset in the political domain collected during the 2020 U.S. elections. Conforti \etal~\cite{Alpher29} constructed WT-WT, a financial dataset containing tweets and annotations carried out by experts. Mohammad \etal~\cite{Alpher01} presented the Stance Dataset consisting of target pairs annotated for the stance of tweeters toward the target.

We present MGTAB, the first stance detection dataset with user network graphs. The large-scale and high-quality annotation of MGTAB will facilitate the development of user stance detection. Additionally, MGTAB provides opportunities for studying graph-based approaches in stance detection.

\begin{table}[htbp]
  \centering
    \scalebox{0.8}{
\begin{tabular}{|l|r|c|c|c|}
\hline
\multirow{3}{*}{Dataset} &\multirow{3}{*}{Samples}& \multicolumn{2}{c|}{Annotation} & \multirow{3}{*}{Graph} \\
\cline{3-4}
\multicolumn{1}{|c|}{} & & \multirow{2}{*}[-0.1ex]{Instance}  & \multirow{2}{*}[0.8ex]{Expert-} &  \\
& &  & \multirow{1}{*}[0.1ex]{annotated} & \\
\hline
\hline
SemEval-2016 T6 & 4,870 & tweet &\XSolidBrush  &\XSolidBrush \\
\hline
SemEval-2019 T7 & 7,730 & tweet &\XSolidBrush  &\XSolidBrush \\
\hline
COVIDLies & 8,937 & tweet &\textcolor{red}\CheckmarkBold  &\XSolidBrush \\
\hline
COVID-19-Stance & 7,122 & tweet &\XSolidBrush  &\XSolidBrush \\
\hline
COVMis-Stance & 2,631 & tweet &\XSolidBrush  &\XSolidBrush\\
\hline
WT-WT & 51,284 & tweet &\textcolor{red}\CheckmarkBold  &\XSolidBrush \\
\hline
P-STANCE & 21,574 & tweet &\XSolidBrush  &\XSolidBrush\\
\hline
Stance Dataset & 4,870 & tweeter &\XSolidBrush  &\XSolidBrush\\
\hline
\textbf{MGTAB} (ours) & 410,199 & tweeter &\textcolor{red}\CheckmarkBold  &\textcolor{red}\CheckmarkBold \\
\hline
\end{tabular}}
\vspace{0.2cm}
\caption{Statistics about our benchmark versus existing stance detection datasets. Compared to other datasets, MGTAB explicitly provide the graph structure among users.}
\vspace{-0.4cm}
\label{tab:1}%
\end{table}%

\subsection{Bot Detection}
The existing bot detection methods can be divided into feature-based methods and graph-based methods.

\noindent
\textbf{Feature-based methods.} Feature-based methods involve the extraction and design of features from user metadata, followed by the use of traditional classifiers for bot detection. Early works~\cite{Alpher10,Alpher32} used simple features such as followers count, friends count, tweets count, and creation date, etc. Some studies have used more complex features, such as features based on social relationships~\cite{Alpher07,Alpher33}. There are also research using the features of user tweets~\cite{Alpher32,Alpher34}. For extracted user features, many studies~\cite{Alpher34,Alpher36,Alpher37,Alpher38,Alpher39} use machine learning algorithms for bot detection. Adaboost (AB)~\cite{Alpher57}, Random Forest (RF)~\cite{Alpher59}, Decision Tree (DT)~\cite{Alpher60}, and SVM~\cite{Alpher61} have all been applied to bot detection. However, the bot may change the registration information according to the features designed for detection to evade feature-based detection methods~\cite{Alpher04,Alpher12}.

\noindent
\textbf{Graph-based methods.}
Graph-based methods are generally more effective than feature-based methods~\cite{Alpher12}. SATAR~\cite{Alpher40} is constructed based on the social graph of Twitter users using feature engineering. GNNs could extract latent representation from complex relations. Inspired by the success of GNNs, Alhosseini \etal~\cite{Alpher41} first attempt to use Graph Convolutional Neural Networks (GCN)~\cite{Alpher42} for spam bot detection that efficiently exploits the graphical structure and relationships of Twitter accounts. Guo \etal~\cite{Alpher43} symmetrically combine BERT and GCN, utilizing text and graph-based features. Some recent studies~\cite{Alpher06,Alpher44,Alpher45,Alpher46} investigate multiple relationships in social graphs. BotRGCN~\cite{Alpher06} constructs a heterogeneous graph through a user network and applies a relational graph convolutional network to bot detection. RGT~\cite{Alpher44} uses relational graph transformers to model the interaction between users in the heterogeneous social graph. However, limited by the lack of relationships in bot detection datasets, the previous researches have used only two types of relationships, friend and follower. The use of multiple relationships in social graphs for bot detection remains largely unexplored.

\noindent
\textbf{Bot detection dataset.}
Despite the highest quality of expert annotation, only the Varol-icwsm is fully annotated by experts due to high costs. Most of the datasets are annotated by crowdsourcing, while others are created using automated techniques based on account behavior, filters on metadata, or others more sophisticated procedures. We summarize the existing bot detection datasets, as shown in Tab.~\ref{tab:2}.

\begin{table}[htbp]
  \centering
    \scalebox{0.8}{
    \begin{tabular}{|l|r|c|c|c|c|}
    \hline
    \multirow{2}{*}{Dataset}  & \multicolumn{1}{c|}{\multirow{2}{*}{Users}}  &\multirow{2}{*}{Semantic} & \multirow{2}{*}[0.8ex]{Expert-}   &\multirow{2}{*}{Graph} \\
    &  & & \multirow{1}{*}[0.1ex]{annotated}  &\\
    \hline
    \hline
    Caverlee    & 30,316   &\textcolor{red}\CheckmarkBold  &\XSolidBrush &\XSolidBrush  \\
    \hline
    Varol-icwsm  & 2,228   &\XSolidBrush &\textcolor{red}\CheckmarkBold &\XSolidBrush  \\
    \hline
    Gilani-17    & 2,484    &\XSolidBrush&\XSolidBrush &\XSolidBrush  \\
    \hline
    Midterm-18   & 50,538   &\XSolidBrush &\XSolidBrush &\XSolidBrush  \\
    \hline
    Cresci-stock  & 13,276  &\XSolidBrush &\XSolidBrush &\XSolidBrush  \\
    \hline
    Cresci-rtbust  & 693    &\XSolidBrush &\XSolidBrush &\XSolidBrush  \\
    \hline
    Botometer-feedback & 518 &\XSolidBrush &\XSolidBrush  &\XSolidBrush  \\
    \hline
    Kaiser      & 1,374     &\XSolidBrush &\XSolidBrush  &\XSolidBrush  \\
    \hline
    Cresci-17   & 14,368     &\textcolor{red}\CheckmarkBold &\XSolidBrush &\XSolidBrush   \\
    \hline
    Cresci-15   & 5,301     &\textcolor{red}\CheckmarkBold &\XSolidBrush &\textcolor{red}\CheckmarkBold   \\
    \hline
    TwiBot-20   & 229,580   &\textcolor{red}\CheckmarkBold &\XSolidBrush &\textcolor{red}\CheckmarkBold  \\
    \hline
    TwiBot-22   &1,000,000   &\textcolor{red}\CheckmarkBold &\XSolidBrush &\textcolor{red}\CheckmarkBold  \\
    \hline
    \textbf{MGTAB} (ours)  & 410,199  &\textcolor{red}\CheckmarkBold &\textcolor{red}\CheckmarkBold  &\textcolor{red}\CheckmarkBold  \\
    \hline
  \end{tabular}}
  \vspace{0.2cm}
  \caption{Statistics about our benchmark versus existing bot detection datasets. Semantic represents datasets including tweets.}
  \vspace{-0.4cm}
  \label{tab:2}
\end{table}

The Caverlee~\cite{Alpher47} consists of bot accounts lured by honeypot accounts, verified human accounts, and their most recent tweets. Varol-icwsm~\cite{Alpher03} dataset consists of manually labeled Twitter accounts sampled from different Botometer score deciles~\cite{Alpher48}. In Gilani-17~\cite{Alpher49}, Twitter accounts were grouped into four categories based on the number of followers. Apart from that, Midterm-18~\cite{Alpher50}, Cresci-17~\cite{Alpher04}, Botometer-feedback~\cite{Alpher35}, Cresci-stock~\cite{Alpher51}, Cresci-rtbust~\cite{Alpher52}, Kaiser~\cite{Alpher53} are also bot detection datasets with various annotation methods and information completeness.

Although there are many bot detection datasets, few have graph structure. Only three publicly available bot detection datasets provide social graphs: Cresci-15~\cite{Alpher10}, TwiBot-20~\cite{Alpher11}, and TwiBot-22 ~\cite{Alpher12}. Cresci-15 and TwiBot-20 contain only two types of relationships, friend and follower, making it difficult to support the research of multi-relational graph-based detection. In TwiBot-22, 1,000 manually labeled accounts are used to train models to get the labels of the remaining accounts, resulting in label deviations. Our proposed MGTAB is fully expert-annotated and has 7 types of relationships. Compared with most previous datasets, it has a larger scale, higher quality annotations, and richer relationships. The annotation comparison is presented in Tab.~\ref{tab:A6}.

\section{Dataset Preprocess}
\label{sec:datapre}
\subsection{Data Collection and Cleaning}
\label{sec:datapre-1}
We adopt breadth-first search (BFS) to obtain the user network of MGTAB, which is based on the selection of 100 seed accounts that are significantly engaged in the debate regarding Japan's decision to discharge nuclear wastewater into the ocean, see Sec.~\ref{sec:MGTABDetails} for details. We collected 10,000 most recent tweets for each user, sufficient for the account detection. The collected data contains a total of 1,554,000 users and 135,450,000 tweets.

We first removed the noisy data and outlier nodes to construct a compact graph. Specifically, users without followers or friends were removed. We then discarded users that were not closely relevant to the target online event and eventually preserved 410,199 accounts and more than 40 million tweets. More information about the data cleaning process is presented in Sec.~\ref{sec:MGTABDetails}.

\subsection{Expert Annotation}
\label{sec:datapre-2}
We invited 12 experts in bot detection and stance detection with more than ten years of working experience to annotate the user stance manually and determine if it is a bot. To further improve the annotation quality, each Twitter user was independently labeled by nine annotators, and annotations for all users were obtained using majority voting. The stances were labeled into three classes: neutral, against, and support. The categories were labeled into two types: human and bot. The distribution of the annotation labels is shown in Tab.~\ref{tab:3}. See comparison of annotations with other datasets in Sec.~\ref{sec:MGTABDetails}. Following TwiBot-20, we use the remaining 400,000 unlabeled users as the support set for research on semi-supervised learning methods.

\begin{table}[htbp]
  \centering
  \caption{Distribution of Labels in annotations.}
  \vspace{0.2cm}
  \scalebox{0.9}{
  \begin{tabular}{|p{3em}|l|c|p{3em}|l|c|}
  \hline
  \multicolumn{3}{|c|}{Stance} & \multicolumn{3}{c|}{Bot} \\
  \hline
  \multicolumn{1}{|p{3em}|}{Label} & \multicolumn{1}{p{3em}|}{Count} & \% & \multicolumn{1}{p{3em}|}{Label} & \multicolumn{1}{p{3em}|}{Count} & \% \\
  \hline
  \hline
  neutral & 3,776 & 37.02 & human & 7,451 & 73.06 \\
  against & 3,637 & 35.66 & bot   & 2,748 & 26.94 \\
  support & 2,786 & 27.32 &       &       & \\
  \hline
  \end{tabular}}
\label{tab:3}
\end{table}

\subsection{Quality Assessment}
\label{sec:datapre-3}
The remaining three experts independently randomly selected 10\% of users labeled to evaluate annotation quality. We obtained 95.4\% accuracy of stance and 97.8\% accuracy of bots on average. This is well above the accuracy obtained in previously released stance detection datasets where crowd-sourcing was used (the accuracy reported, in percentage, ranges from 63.7\% to 79.7\%)~\cite{Alpher29}. In addition, compared to the 80\% and 90.5\% accuracy in TwiBot-20~\cite{Alpher11} and TwiBot-22~\cite{Alpher12}, our 97.8\% accuracy of bots has considerably improved annotation quality.

\subsection{Feature Analysis}
\label{sec:datapre-4}
We randomly selected 2000 labeled users to analyze the effectiveness of features for detection. We analyzed features in different aspects, including creation time, friend count, name length, etc. Following~\cite{Alpher10}, we use the information gain (IG) to measure the informativeness of a feature to the predicting class. It can be informally defined as the expected reduction in entropy caused by the knowledge of the value of a given attribute.

Use $Y$ to denote the user's category, $H(Y)$ to represent the entropy of $Y$, and $y$ is the value of $Y$, $y \in\left\{y_{1}, y_{2}, \ldots, y_{K}\right\}$. In stance detection, $K$ is 3, and in bot detection, $K$ is 2.

\begin{equation}
\label{equ:1}
H(Y)=-\sum_{k=1}^{K} p\left(y_{k}\right) \log _{2} p\left(y_{k}\right).
\end{equation}

$H(Y \mid X)$ denotes $H(Y)$ when the feature $X$ is given and it can be computed by:

\begin{equation}
\label{equ:2}
\vspace{-0.2cm}
H(Y \mid X)=-\sum_{x \in \Phi} p_{x} \sum_{k=1}^{K} p\left(y_{k} \mid x\right) \log _{2} p\left(y_{i} \mid x\right),
\end{equation}

where $x$ is the value of $X$, $x \in \Phi$. The $IG(X ; Y)$ indicates that the category information increases (uncertainty decreases) after $Y$ gets feature $X$:

\begin{equation}
\label{equ:3}
IG(X ; Y)=H(Y)-H(Y \mid X).
\end{equation}

The features with greater IG contain more information for detection. According to the type of features, we divide the features into boolean and numeric features, and the boolean features take the value of True or False. The numeric features are taken the logarithm except for the creation time. Then, the data is divided into $K$ intervals uniformly according to the value domain, the number of samples in each interval is counted, and then the IG is calculated using the discrete values. In this paper, $K$ is set to 51.

\begin{figure*}[t]
  \centering
   \includegraphics[width=0.9\linewidth]{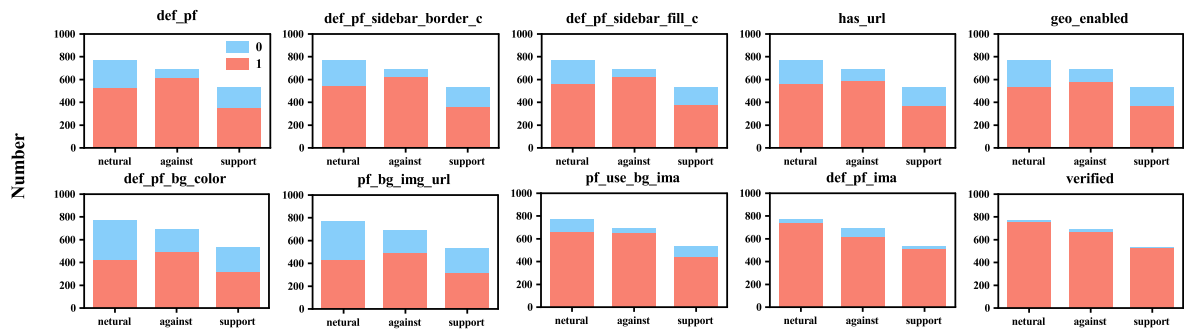}
   \vspace{-0.1cm}
   \caption{Distribution of boolean features with top 10 IG in stance detection.}
   \label{fig:1}
\end{figure*}

\begin{figure*}[t]
  \centering
   \includegraphics[width=0.9\linewidth]{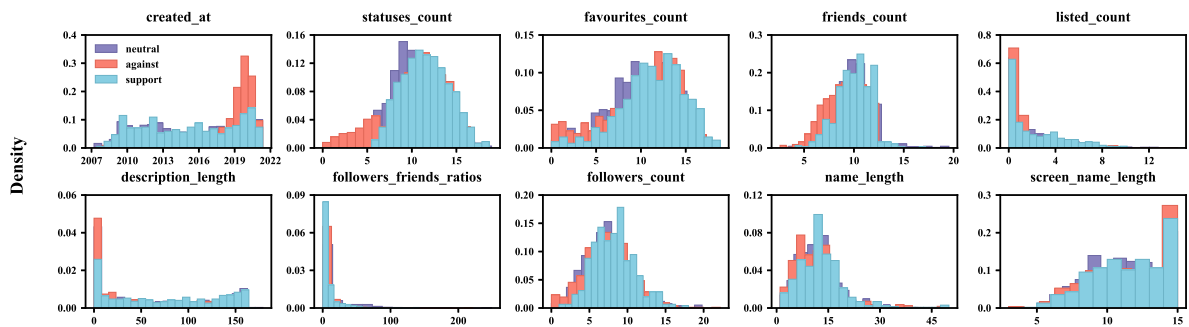}
   \vspace{-0.1cm}
   \caption{Distribution of numerical features with top 10 IG in stance detection.}
   \label{fig:2}
\end{figure*}

\noindent
\textbf{User stance features.}
The features with the same distribution are first removed, and then the IG of the user's features is calculated to obtain the boolean and numerical features with the top 10 IG for bot detection. The boolean and numerical features are shown in Fig.~\ref{fig:1} and~\ref{fig:2}, respectively, in decreasing order of IG.

The boolean and numerical features with the top 3 IG were analyzed: \emph{default\_profile}: Most users with opposing stances prefer to use the default profile. \emph{default\_profile\_sidebar\_border\_color}: Most users with opposing stances prefer to use the default profile's sidebar border color. \emph{default\_profile\_sidebar\_fill\_color}: Most users with opposing stances prefer to use the default profile's sidebar color. \emph{created\_at}: Most users with opposing stances have been created recently. \emph{statues\_count}: Users with opposing stances have a larger share of users with lower statuses. \emph{favourites\_count}: Among the users with lower favourites, those who are opposed are more.

\vspace{-0.1cm}

\begin{figure*}[t]
  \centering
   \includegraphics[width=0.9\linewidth]{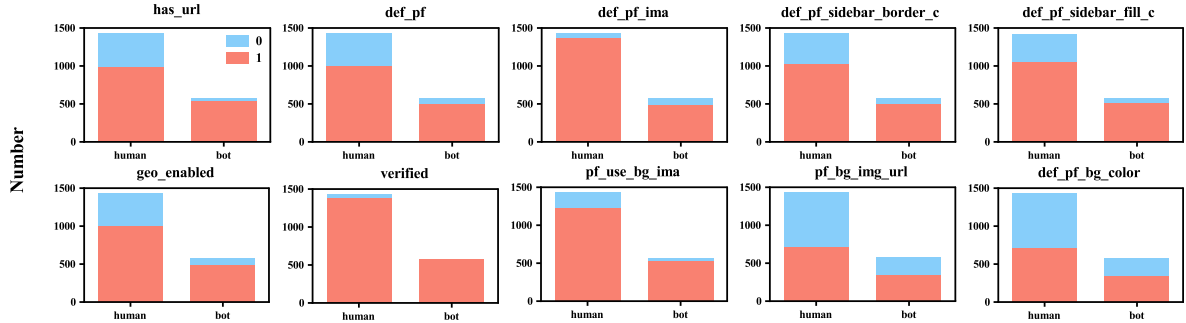}
   \vspace{-0.1cm}
   \caption{Distribution of boolean features with top 10 IG in bot detection.}
   \label{fig:3}
\end{figure*}

\begin{figure*}[t]
  \centering
   \includegraphics[width=0.9\linewidth]{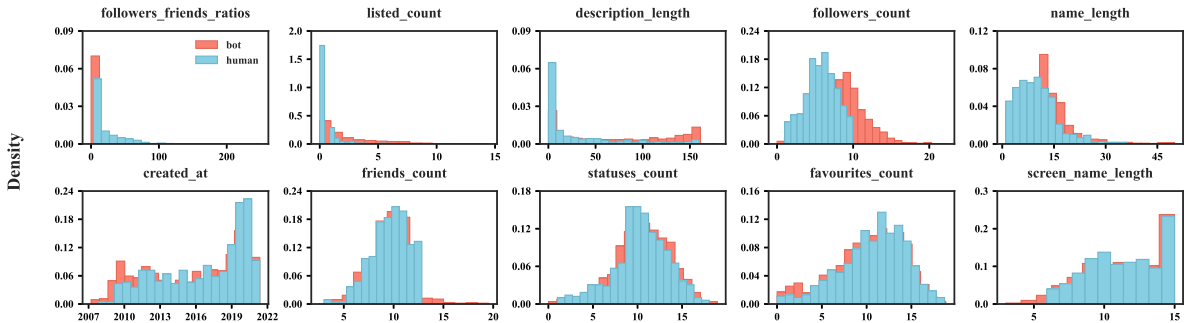}
   \vspace{-0.1cm}
   \caption{Distribution of numerical features with top 10 IG in bot detection.}
   \label{fig:4}
\end{figure*}

\noindent
\textbf{User bot features.}
Conducting the same processing above to obtain the boolean and numerical features with the top 10 IG for stance detection. The boolean and numerical features are shown in Fig.~\ref{fig:3} and~\ref{fig:4}, respectively, in decreasing order of IG.

The boolean and numerical features with the top 3 IG were analyzed: \emph{has\_url}: Most bots have empty URL content. \emph{default\_profile}: Compared to humans, bots tend to use the default profile. \emph{default\_profile\_image}: Most of the users with the default background image are bots. \emph{followers\_friends\_ratios}: Bots usually increase the follower count by following each other, which leads to a smaller followers\_friends ratio. \emph{listed\_count}: Bots belong to more public lists than human users. \emph{description\_length}: In order to masquerade as a human user, bots tend to fill in the account description more often and with longer descriptions than humans.

Our experiments show that the features selected are more effective than those extracted in previous literature~\cite{Alpher06,Alpher39,Alpher50}, the details are presented in Sec.~\ref{sec:FeatureAnalysis}.

\section{Dataset Construction}
\subsection{Feature Representation Construction}
\noindent
We concatenate user property features and user tweet features to serve as user feature representations, $r=\left[r_{prop} \| r_{tweet}\right]$. The details of the user feature representations are shown in Tab.~\ref{tab:4}.

\noindent
\textbf{Property features extraction.}
User property features are obtained based on the analysis in Sec.~\ref{sec:datapre-4}. The selected numerical features are normalized by MinMaxScaler method to obtain the representation of numerical feature $r_{num}$. The selected boolean features are numericalized, where True and False are replaced with 1 and 0, respectively, to obtain the representation of boolean feature $r_{bool}$. The representation of user property features is obtained by concatenating $r_{num}$ and $r_{bool}$, $r_{prop}=\left[r_{num} \| r_{bool}\right]$.

\begin{figure}[h]
  \centering
   \includegraphics[width=0.85\linewidth]{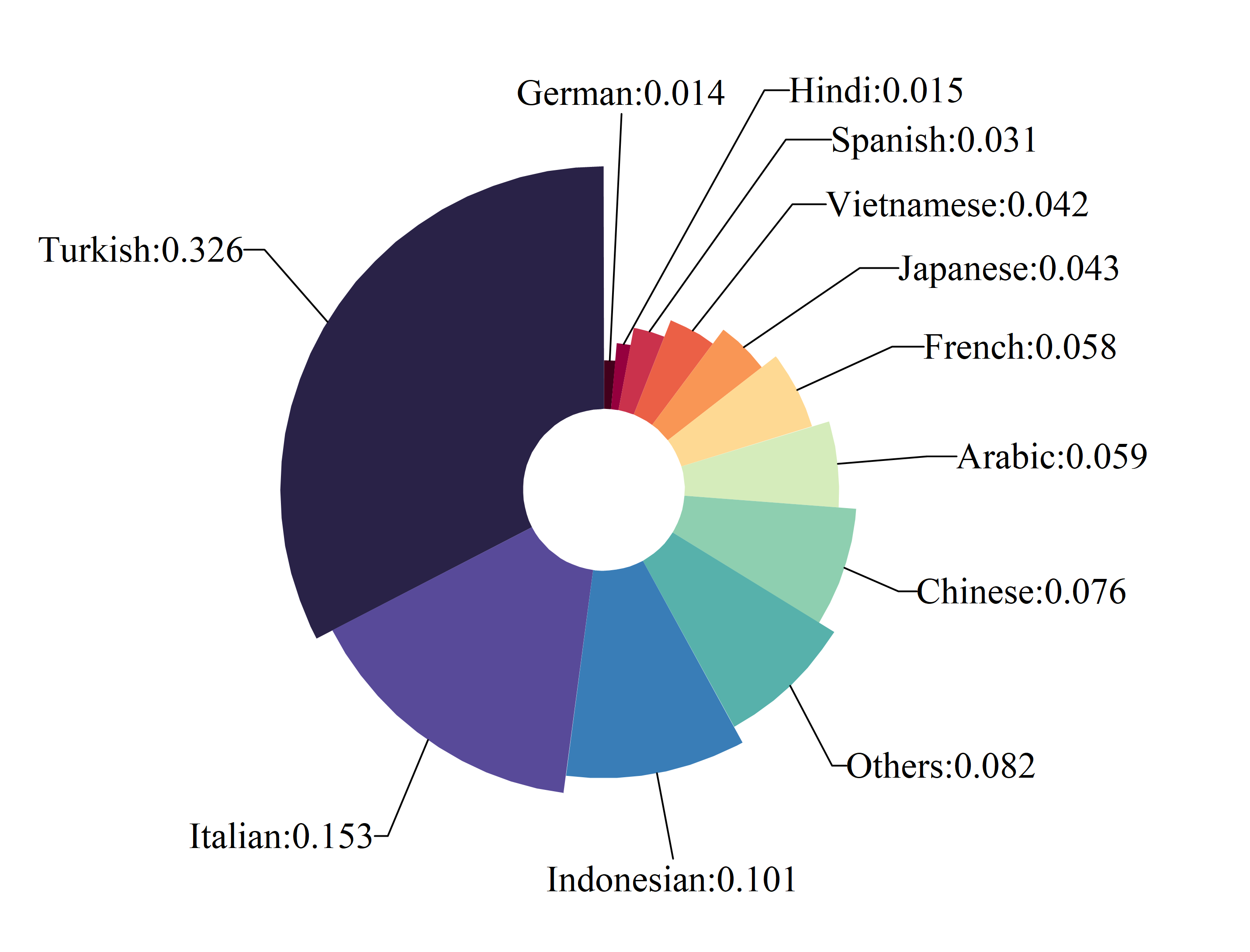}
     \vspace{-0.2cm}
   \caption{Non-English tweets and their percentage.}
     \vspace{-0.2cm}
   \label{fig:5}
\end{figure}

\noindent
\textbf{Tweet features extraction.}
The tweets contain 54 languages, of which English is the most frequent, with a ratio of 73.6\%. More details are available in Sec.~\ref{sec:FeatureAnalysis}, and the statistics of non-English languages are shown in Fig.~\ref{fig:5}. It is not easy to encode multilingual tweets well using a monolingual pre-trained BERT model. Therefore, we use LaBSE~\cite{Alpher54}, a multilingual BERT, to extract tweet features. Specifically, We use LaBSE to encode user tweets. We average the representation of all tweets to obtain the representation of user tweets $r_{tweet}$.
The demonstration of the effectiveness encoded by LaBSE is shown in Sec.~\ref{sec:BERTModels}.

\subsection{Relationship graph construction}
\begin{figure*}[ht]
  \centering
   \includegraphics[width=0.85\linewidth]{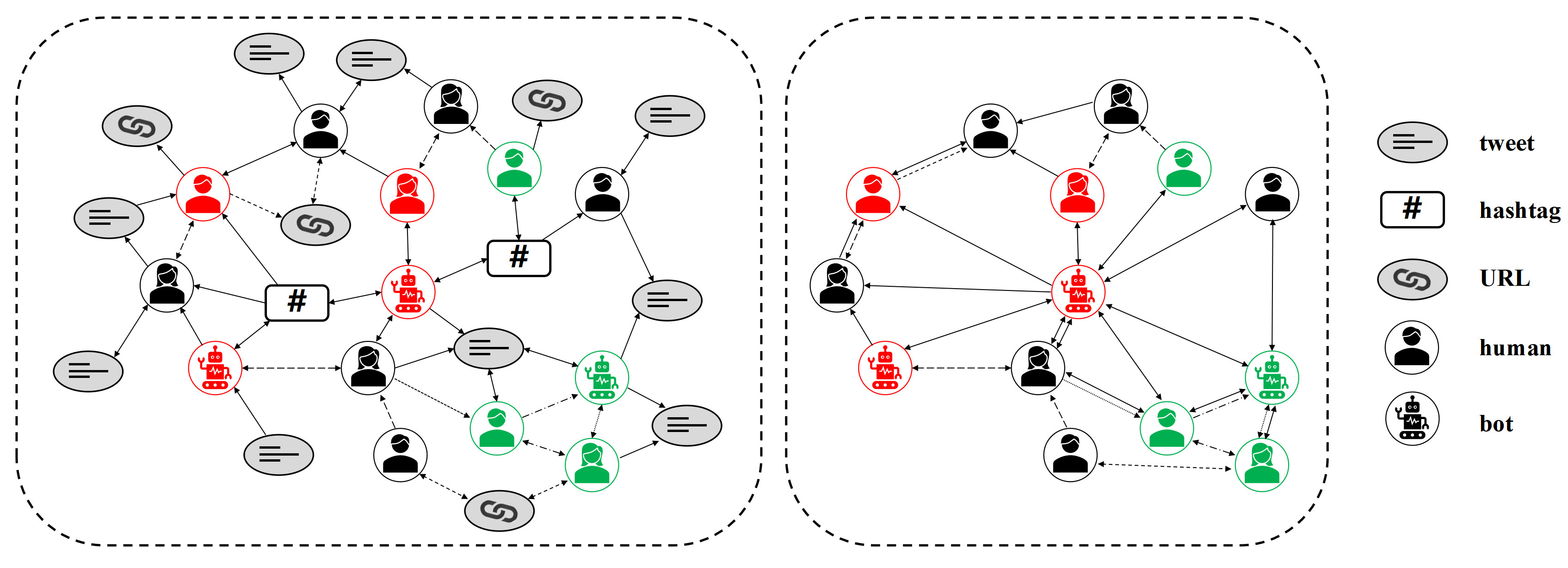}
     \vspace{-0.2cm}
   \caption{We simplify the original complex heterogeneous graph network (left) and construct a user-level multi-graph network (right). Black, red, and green denote neutral, against, and support.}
     \vspace{-0.2cm}
   \label{fig:7}
\end{figure*}

The complex social graph structure, including multiple entities such as users, tweets, hashtags, URLs, etc., makes graph-based account detection a complex problem. Since the focus of attention in user-level detection is on the user. The recently proposed state-of-the-art detection methods based on heterogeneous graphs ~\cite{Alpher06,Alpher44,Alpher45,Alpher46} only use the relationship between users. Therefore, to simplify the social network graph, we only kept users as nodes when constructing the social graph, as shown in Fig.~\ref{fig:7}. For the other types of entities, only the relationships between users are used.

\noindent
\textbf{Explicit relationship extraction.}
For explicit relationships such as follower, friend, mention, reply, and quoted, connections between users are constructed directly from their relationships. The edges constructed based on the above relationships are all directed edges, as shown in Tab.~\ref{tab:5}.

\noindent
\textbf{Implicit relationship construction.}
We also extracted 2 implicit relationships between users: URL co-occurrence and hashtag co-occurrence. Specially, the co-occurrences relationship between user nodes ${v}_{i}$ and ${v}_{j}$ can be determined by the probability of entities co-occurring, whose weight is calculated through average Pointwise Mutual Information (PMI):

\vspace{-0.2cm}
\begin{equation}
\label{equ:4}
\mathrm{W}\left(v_{i}, v_{j}\right)=\frac{1}{\left|\Psi_{\{i, j\}}\right|} \sum_{e_{k} \in \Psi_{\{i, j\}}} \log \frac{p\left(v_{i}, e_{k}\right) p\left(v_{j}, e_{k}\right)}{p\left(e_{k}\right)^{2}},
\end{equation}

where $\Psi_{\{i, i\}}$ denotes the set of entities common to ${v}_{i}$ and ${v}_{j}$. Use $1/N_i$ approximates $p\left(v_{i}, e_{k}\right)$ when calculate PMI, where $N_i$ denote the length of entities list of ${v}_{i}$. Finally, we obtain the MGTAB heterogeneous graphs containing 410,199 nodes and over 100 million edges.

\begin{table}[htbp]
  \centering
  \caption{Relations in the MGTAB heterogeneous graph.}
  \vspace{0.2cm}
  \scalebox{0.80}{
    \begin{tabular}{|p{3em}|p{2.65em}|p{2.65em}|p{16.25em}|}
    \hline
    \multirow{2}{*}{Relation} & \multicolumn{2}{c|}{Direction} & \multirow{2}{*}{Description}\\
    \cline{2-3}
    \multicolumn{1}{|l|}{} & Source & \multicolumn{1}{p{2.65em}|}{Target} & \multicolumn{1}{l|}{}\\
    \hline
    \hline
    follower & user A & \multicolumn{1}{p{2.65em}|}{user B} & user A is followed by user B \\
    friend & user A & \multicolumn{1}{p{2.65em}|}{user B} & user A follows user B \\
    mention & user A & \multicolumn{1}{p{2.65em}|}{user B} & user A mentions user B in tweets \\
    reply & user A & \multicolumn{1}{p{2.65em}|}{user B} & user A replies to tweet of user B \\
    quote & user A & \multicolumn{1}{p{2.65em}|}{user B} & user A quotes tweet of user B \\
    \hline
    URL   & \multicolumn{2}{c|}{Undirected} & user A and user B have the same URL \\
    hashtag & \multicolumn{2}{c|}{Undirected} & user A and user B have the same hashtag \\
    \hline
    \end{tabular}}
  \label{tab:5}%
\end{table}%

\section{Experiments}
\subsection{Experiment Settings}

\noindent
\textbf{Datasets.}
In stance detection, we evaluate models on our proposed benchmark, SemEval-2016 T6~\cite{Alpher21}, and SemEval-2019 T7~\cite{Alpher25}. In bot detection, in addition to our proposed benchmark, we evaluate models on 4 publicly available bot detection datasets: Cresci-17~\cite{Alpher04}, Cresci-15~\cite{Alpher10}, TwiBot-20~\cite{Alpher11}, and TwiBot-22~\cite{Alpher12}. We use all the annotated data in experiments. Following~\cite{Alpher11,Alpher12}, we conduct a 7:2:1 random partition as training, validation, and test set for all datasets.

\noindent
\textbf{Baselines.}
We use competitive and state-of-the-art stance dection and bot detection methods include: Adaboost Classifier (AB) ~\cite{Alpher57}, Decision Tree (DT)~\cite{Alpher60}, Random Forest (RF)~\cite{Alpher59}, Support Vector Machines (SVM)~\cite{Alpher61}, Graph Convolutional Network (GCN)~\cite{Alpher42}, Graph Attention Network (GAT)~\cite{Alpher20}, Heterogeneous Graph Transformer (HGT)~\cite{Alpher55}, Simple Heterogeneous Graph Neural Network (S-HGN)~\cite{Alpher56}, Bot Detection with Relational Graph Convolutional Networks (BotRGCN)~\cite{Alpher06}, and Relational Graph Transformers (RGT)~\cite{Alpher44}.

\subsection{Benchmark Performance}
\label{sec:benchmarkPerformance}
We evaluate baselines on datasets and present their detection accuracy and F1-score in Tab.~\ref{tab:6}. All hyper-parameters are listed in Sec.~\ref{sec:ComputationDetails} for replication.

We observed that the graph-based methods performed better than feature-based methods, all top 3 models are graph-based. In addition, it is obvious to observe that heterogeneous GNNs perform better than homogeneous GNNs. We speculate that this is because heterogeneous GNNs are sufficient to capture the multiple relationships between users.
RGT could model the heterogeneous influence between users, achieving the best performance on most datasets. Better utilizing weights and directions of the edge is a potential future research direction.

\begin{table*}[htbp]
  \centering
  \caption{
  Performance of baseline methods on datasets. Use the most commonly used follower and friend relationships during evaluation. Each baseline is conducted five times with different seeds, and we report the average performance and standard deviation. ``/'' indicates that the dataset does not contain user relationships to support the grah-based methods. Best and second best results are highlighted in \textbf{bold} and \underline{underline}.}
    \vspace{0.2cm}
  \scalebox{0.76}{
\begin{tabular}{|c|l|c|c|c|c|c|c|c|c|c|c|c|}
\hline
\multirow{4}{*}{Task} & \multirow{4}{*}{Dataset} & \multirow{4}{*}{Metric} & \multicolumn{10}{c|}{Methods} \\
\cline{4-13}
&   &  & \multicolumn{4}{c|}{\multirow{2}{*}[1.25ex]{Feature-based}} & \multicolumn{6}{c|}{Graph-based} \\
\cline{8-13}
&  &  &   \multicolumn{4}{c|}{}  &\multicolumn{2}{c|}{Homogeneous} & \multicolumn{4}{c|}{Heterogeneous} \\
\cline{4-13}
&  &  & AB  & RF  & DT  & SVM & GCN & GAT   & HGT   & S-HGN & BotRGCN & RGT\\
\hline
\hline
\multirow{6}{*}{Stance} & \multirow{2}{*}{SemEval-2016} & Acc   & \underline{74.2±0.4} & 72.8±0.3 & 72.2±0.3 & \textbf{76.1±0.3} & / & / & / & / & / & / \\
\cline{3-13}
&       & F1    & \underline{72.1±0.3} & 70.3±0.3 & 69.2±0.3 & \textbf{72.3±0.3} & / & / & / & / & / & / \\
\cline{2-13}
& \multirow{2}{*}{SemEval-2019} & Acc   & \underline{84.0±0.5} & 83.8±0.4 & 83.1±0.4 & \textbf{84.2±0.4} & / & / & / & / & / & / \\
\cline{3-13}
&       & F1    & \underline{64.4±0.4} & 63.6±0.3 & 64.1±0.4 & \textbf{65.2±0.4} & / & / & / & / & / & / \\
\cline{2-13}
& \multirow{2}{*}{MGTAB} & Acc & 74.6±1.4 & 79.6±0.7 & 66.9±0.9 & 81.2±0.7 & 82.4±0.9 & 82.2±1.2 & 83.2±0.4 & \underline{85.3±0.5} & 84.7±1.5 & \textbf{87.8±0.4} \\
\cline{3-13}  &   & F1 & 73.9±1.5 & 79.0±0.8 & 66.0±0.8 & 80.7±0.8 & 81.5±0.9 & 81.0±1.2 & 81.8±0.3 & \underline{84.4±0.4} & 84.3±1.4 & \textbf{86.9±0.4}\\
\hline
\multirow{10}{*}{Bot} &\multirow{2}{*}{Cresci-17} & Acc  & \textbf{91.2±0.2} & \underline{89.1±0.2} & 86.2±0.2 & 84.1±0.3 & / & / & / &/ & / & / \\
\cline{3-13}
&   & F1  & \textbf{83.4±0.2} & \underline{80.9±0.2} & 76.4±0.2 & 72.8±0.3 & /  & /  & /  & / & /  & /\\
\cline{2-13}
& \multirow{2}{*}{Cresci-15} & Acc   & 95.9±0.3 & 97.0±0.8 & 96.2±1.3 & 96.6±0.2 & 98.2±0.6 & 98.1±0.2 & 98.4±0.3 & 97.5±0.5 & \underline{98.5±0.4} & \textbf{98.6±0.3} \\
\cline{3-13}
&    & F1 & 95.5±0.3 & 96.7±0.9 & 95.9±1.4 & 96.3±0.3 & 98.0±0.4 & 98.0±0.1 & 98.4±0.3 & 97.2±0.5 & \underline{97.3±0.5} & \textbf{98.5±0.2}\\
\cline{2-13}
& \multirow{2}{*}{TwiBot-20} & Acc   & 85.7±0.4 & 85.0±0.5 & 80.1±0.5 & 85.2±0.3 & 77.2±1.2 & 83.2±0.4 & 85.9±0.6 & 85.4±0.3 & \underline{86.8±0.5} & \textbf{86.9±0.3} \\
\cline{3-13}
&    & F1 & 85.6±0.4 & 84.9±0.5 & 80.0±0.5 & 84.8±0.4 & 76.6±0.4 & 81.9±0.5 & 85.6±0.6 & 85.3±0.2 & \underline{86.6±0.4} & \textbf{86.7±0.4} \\
\cline{2-13}
& \multirow{2}{*}{TwiBot-22} & Acc   & 69.3±0.5 & 74.3±0.7 & 72.6±0.8 & 76.4±0.9 & 78.3±1.3 & \underline{79.3±0.8} & 74.9±1.2 & 76.7±1.3 & \textbf{79.6±0.4} & 76.5±0.4 \\
\cline{3-13}
&   & F1 & 34.8±0.5 & 30.4±0.6 & 51.6±0.6 & 54.6±0.8 & 54.8±1.0 & \underline{55.6±1.1} & 39.2±1.6 & 45.7±0.5 & \textbf{57.6±1.4} & 43.1±0.5\\
\cline{2-13}
& \multirow{2}{*}{MGTAB} & Acc   & 90.1±0.9 & 89.5±0.4 & 87.1±0.5 & 88.7+1.4 & 85.8±1.3 & 87.0±1.3 & 90.3±0.3 & \underline{91.4+0.4} & 89.6±0.8 & \textbf{92.1+0.4}\\
\cline{3-13}
&   & F1 & 87.7±1.1 & 86.8±0.5 & 83.7±0.7 & 85.3+1.7 & 78.3±1.7 & 82.3±2.1 & 87.5±0.4 & \underline{88.7+0.6} & 87.2±0.7 & \textbf{90.4±0.5} \\
\hline
\end{tabular}}
  \label{tab:6}%
\end{table*}

\begin{table*}[htbp]
  \centering
  \caption{Accuracy of graph-based detection methods on MGTAB using different relations. Each baseline is conducted five times with different seeds, and we report the average performance and standard deviation. Best results are highlighted in \textbf{bold}.}
    \vspace{0.2cm}
\scalebox{0.76}{
\begin{tabular}{|c|l|c|c|c|c|c|c|c|c|c|c|}
\hline
\multirow{3}{*}{Task} & \multirow{3}{*}{Method} & \multicolumn{10}{c|}{Relation} \\
\cline{3-12}      &       & \multicolumn{7}{c|}{Single relation}                  & \multicolumn{3}{c|}{Multiple relations} \\
\cline{3-12}      &       & \multicolumn{1}{c|}{1:follower} & \multicolumn{1}{c|}{2:friend} & \multicolumn{1}{c|}{3:mention} & \multicolumn{1}{c|}{4:reply} & \multicolumn{1}{c|}{5:quoted} & \multicolumn{1}{c|}{6:hashtag} & \multicolumn{1}{c|}{7:url} & \multicolumn{1}{c|}{1+2} & \multicolumn{1}{c|}{3+4+5} & \multicolumn{1}{c|}{1+2+3+4+5+6} \\
\hline
\hline
\multirow{5}{*}{stance} & GCN   & 76.7±0.6 & 76.9±0.6 & 76.8±0.7 & 75.9±1.0 & 76.9±0.6 & 60.8±1.2 & 76.2±0.4 & 78.1±0.6 & 77.1±0.5 & \textbf{79.1±0.3} \\
\cline{2-12}      & GAT   & 77.0±0.5 & 76.7±0.5 & 78.0±0.4 & 76.8±0.2 & 77.3±0.3 & 64.4±1.1 & 77.1±0.4 & 77.8±0.4 & 77.3±0.4 & \textbf{77.9±0.4}\\
\cline{2-12}      & BotRGCN & 79.1±0.3 & 76.1±0.4 & 76.2±0.5 & 76.8±0.6 & 77.0±0.2 & 77.2±0.2 & 76.8±0.5 & 79.4±0.2 & 78.0±0.4 & \textbf{79.2±0.5} \\
\cline{2-12}      & S-HGN & 81.2±0.2 & 80.8±0.2 & 79.4±0.2 & 78.5±0.2 & 78.6±0.3 & 75.6±0.3 & 80.3±0.3 & 81.6±0.2 & 80.0±0.2 & \textbf{81.7±0.2} \\
\cline{2-12}      & HGT   & 79.1±0.1 & \textbf{79.6±0.2} & 77.4±0.2 & 77.4±0.2 & 77.7±0.2 & 78.1±0.3 & 77.0±0.2 & 79.2±0.2 & 77.9±0.2 & 78.7±0.1\\
\hline
\multirow{5}{*}{bot} & GCN   & 81.2±0.5 & 84.1±0.7 & 84.6±0.3 & 85.2±0.6 & 85.5±0.6 & 76.3±1.2 & 81.5±0.4 & 83.6±0.5 & \textbf{86.2±0.4} & 82.5±0.5 \\
\cline{2-12}      & GAT   & 81.2±1.5 & 83.0±1.6 & 83.3±2.0 & 83.2±0.2 & 82.9±0.3 & 73.6±1.6 & 79.7±0.4 & 84.4±0.8 & \textbf{86.4±0.3} & 78.4±0.9 \\
\cline{2-12}      & BotRGCN & 83.5±0.5 & 83.2±0.3 & 82.9±0.2 & 83.2±0.3 & 82.9±0.5 & 83.1±0.3 & 82.4±0.3 & 86.9±0.2 & 86.6±0.3 & \textbf{87.2±0.2} \\
\cline{2-12}      & S-HGN & 87.5±0.3 & 87.3±0.3 & 87.3±0.3 & 86.8±0.3 & 86.9±0.2 & 86.9±0.3 & 86.8±0.2 & 87.3±0.4 & 87.4±0.3 & \textbf{87.9±0.2} \\
\cline{2-12}      & HGT   & 87.1±0.3 & 87.4±0.4 & 86.5±0.4 & 86.4±0.4 & 86.6±0.3 & 85.6±0.3 & 85.7±0.2 & \textbf{87.5±0.2} & 86.4±0.0 & 87.2±0.1 \\
\hline
\end{tabular}}
\label{tab:7}%
\end{table*}%

\subsection{Study of Training Set Size}
\begin{figure}[h]
\centering
\vspace{-0.2cm}
\centerline{\includegraphics[width=0.9\linewidth]{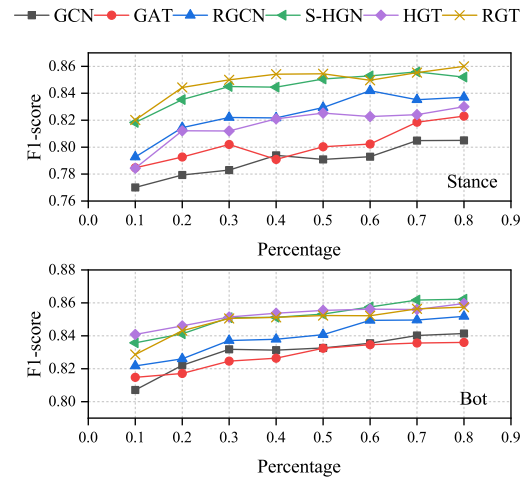}}
\vspace{-0.2cm}
\caption{Performance of graph-based methods in stance detection (above) and bot detection (below) when trained on different percentages of labeled users on MGTAB.}
\label{fig:6}
\end{figure}

We select each 10\% of the labeled users as the test and validation sets. Then, we utilize different proportions of labeled users as the training set, increasing from 10\% to 80\%. The graph-based model performances under different training sets are shown in Fig.~\ref{fig:6}.

The heterogeneous GNNs' performance is better than homogeneous GNNs under different training sets. This phenomenon is consistent with the results in Sec.~\ref{sec:benchmarkPerformance}.

As more annotated data is used, all detection models become more effective. Existing account detection methods are generally supervised and rely on large amounts of labeled data. MGTAB's large scale contributes to training better detection models. In addition, MGTAB provides 400,000 unlabeled users to support the study of semi-supervised account detection methods. To the best of our knowledge, MGTAB has the most unlabeled users in the account detection field.

\subsection{Social Graph Relationship Analysis}
In this section, we analyze the impact of using various relationships in the MGTAB. In addition to single relationships, we also experimented with using multiple relationships. We randomly conduct a 1:1:8 partition as training, validation, and test set. This partition is shared across all experiments in Sec.~\ref{sec:FeatureAnalysis} and~\ref{sec:BERTModels}.

Tab.~\ref{tab:7} illustrates that graph-based account detection methods perform better when more relationships are used. This trend suggests that future research in account detection should focus on better utilizing multiple relationships between users. Besides, we observed that hashtag co-occurrence has the worst performance of all the relationships. We suspect this is because hashtag co-occurrence is highly random, and two unrelated users can have hashtag co-occurrence. Although MGTAB provides edges for URL and hashtag co-occurrence relationships, existing graph-based account detection models cannot fully exploit them, leading to bad performance.

\section{Conclusion}
In this paper, we presented MGTAB, a large-scale dataset for stance detection and bot detection. The dataset was created using expert annotation and majority voting to ensure high-quality annotations. To build the normalized dataset, we selected 20 user features with the highest information gain, which was experimentally demonstrated to be the most effective. We extracted 7 types of relationships between users and simplified the complex Twitter network. Compared to previous datasets, MGTAB provides better support for the study of graph-based account detection methods. Our experiments showed that graph-based approaches are generally more effective than feature-based approaches and perform better when multiple relationships are introduced.

{\small
\bibliographystyle{ieee_fullname}
\bibliography{egbib}
}

\clearpage

\appendix
\renewcommand\thetable{\Alph{section}\arabic{table}}
\section{Appendix}
\setcounter{table}{0}
\subsection{MGTAB Details}
\label{sec:MGTABDetails}
\noindent
\textbf{Seed accounts selection.}
We have chosen 100 seed accounts from users actively participating in the discussion regarding Japan's plan to release nuclear wastewater into the sea. These accounts consist of 73 high-influence accounts that have a large number of followers, handpicked by experts. Additionally, 27 accounts were randomly selected to ensure diversity in the sample.

\noindent
\textbf{Expert annotation details.}
The annotation process of the 100 seed accounts involved experts thoroughly checking the personal information and the last 100 tweets of each account. Based on their evaluation, the experts categorized the user's stance towards the discharged and treated nuclear wastewater in Japan as either supportive, neutral, or opposing. Additionally, they classified the user's behavior as either human or bot. The entire dataset was annotated by nine annotators over a period of four months. The average Fleiss' Kappa of stance and bot annotation among the nine annotators was 0.692 and 0.823, respectively. The high average Fleiss' Kappa values suggest that there was good agreement among the annotators in categorizing the user's stance and behavior.

\noindent
\textbf{Data cleaning details.}
Tab.~\ref{tab:A6} displays the comparison of annotations between the datasets that have been manually annotated, as listed in Tab.~\ref{tab:2}. The Varol-icwsm, Gilani-17, Cresci-rtbust-19, Botometer-feedback, and Cresci-17 datasets lack tweet information and relationships between users.

\begin{table}[htbp]
  \centering
  \caption{Distribution of Labels in annotations. }
  \vspace{0.2cm}
  \scalebox{0.85}{
  \begin{tabular}{|c|c|c|}
  \hline
  Dataset & Manual annotations & Total accounts \\
  \hline
  \hline
  Varol-icwsm           & 2,228    & 2,228     \\
  Gilani-17             & 2,484    & 2,484     \\
  Cresci-rtbust-19      & 693      & 693       \\
  Botometer-feedback    & 518      & 518       \\
  Cresci-17             & 11,017   & 14,368    \\
  \hline
  TwiBot-20             & 11,826   & 229,580   \\
  TwiBot-22             & 1,000    & 1,000,000 \\
  \textbf{MGTAB}(ours)  & 10,199   & 410,199   \\
  \hline
\end{tabular}}
\label{tab:A6}
\end{table}

Cresci-2017 utilized a diverse range of labeling techniques, including the procurement of 3,351 spurious followers. In TwiBot-22, a group of one 1000 accounts has been manually labeled and employed to train models that can anticipate the labels of the remaining accounts in the dataset.

\noindent
\textbf{Data cleaning details.}
In order to construct a more compact and relevant graph, users without followers or friends were removed. If a user does not meet any of the following three conditions, they will be also discarded:

\begin{itemize}
\vspace{-0.3cm}
\item They did not participate in the discussion related to "Japan's release of radioactive wastewater".
\vspace{-0.3cm}
\item None of their published tweets contain keywords related to the event
\vspace{-0.3cm}
\item They do not have any friend or follower relationship with other users who participated in the discussion.
\end{itemize}

\begin{table}[htbp]
  \centering
  \caption{Top 10 features for user stance detection, listed in descending order of IG.}
  \vspace{0.2cm}
  \scalebox{0.9}{
    \begin{tabular}{|l|c|c|}
    \hline
     Features & IG    & Type \\
    \hline
    \hline
    def\_pf & 0.044927 & \multirow{10}[1]{*}{Boolean} \\
    \cline{1-2}
    def\_pf\_sidebar\_border\_c & 0.041702 &  \\
    \cline{1-2}
    def\_pf\_sidebar\_fill\_c & 0.035676 &  \\
    \cline{1-2}
    has\_URL & 0.019093 &  \\
    \cline{1-2}
    geo\_enabled & 0.018741 &  \\
    \cline{1-2}
    def\_pf\_bg\_color & 0.016575 &  \\
    \cline{1-2}
    pf\_bg\_img\_URL & 0.016373 &  \\
    \cline{1-2}
    pf\_use\_bg\_ima & 0.013501 &  \\
    \cline{1-2}
    def\_pf\_ima & 0.010923 &  \\
    \cline{1-2}
    verified & 0.001304 &  \\

    \hline
    created\_at & 0.107146 & \multirow{10}[1]{*}{Numerical} \\
    \cline{1-2}
    statuses\_count & 0.101377 &  \\
    \cline{1-2}
    favourites\_count & 0.070515 &  \\
    \cline{1-2}
    friends\_count & 0.064378 &  \\
    \cline{1-2}
    listed\_count & 0.063106 &  \\
    \cline{1-2}
    description\_length & 0.062262 &  \\
    \cline{1-2}
    followers\_friends\_ratios & 0.056947 &  \\
    \cline{1-2}
    followers\_count & 0.055433 &  \\
    \cline{1-2}
    name\_length & 0.054854 &  \\
    \cline{1-2}
    screen\_name\_length & 0.015331 &  \\
    \hline
    \end{tabular}}
  \label{tab:A1}%
\end{table}%

\begin{table}[htbp]
  \centering
  \caption{Top 10 features for bot detection, listed in descending order of IG.}
  \vspace{0.2cm}
  \scalebox{0.9}{
    \begin{tabular}{|l|c|c|}
    \hline
     Features & IG    & Type \\
    \hline
    \hline
    has\_URL & 0.064248 & \multirow{10}[1]{*}{Boolean} \\
    \cline{1-2}
    def\_pf & 0.025997 &  \\
    \cline{1-2}
    def\_pf\_ima & 0.025402 &  \\
    \cline{1-2}
    def\_pf\_sidebar\_border\_c & 0.023105 &  \\
    \cline{1-2}
    def\_pf\_sidebar\_fill\_c & 0.022359 &  \\
    \cline{1-2}
    geo\_enabled & 0.019302 &  \\
    \cline{1-2}
    verified & 0.010902 &  \\
    \cline{1-2}
    pf\_use\_bg\_ima & 0.007877 &  \\
    \cline{1-2}
    pf\_bg\_img\_URL & 0.005923 &  \\
    \cline{1-2}
    def\_pf\_bg\_color & 0.005841 &  \\
    \hline
    followers\_friends\_ratios & 0.391857 & \multirow{10}[1]{*}{Numerical} \\
    \cline{1-2}
    listed\_count & 0.333101 &  \\
    \cline{1-2}
    description\_length & 0.194765 &  \\
    \cline{1-2}
    followers\_count & 0.176186 &  \\
    \cline{1-2}
    name\_length & 0.040335 &  \\
    \cline{1-2}
    created\_at & 0.034079 &  \\
    \cline{1-2}
    friends\_count & 0.031598 &  \\
    \cline{1-2}
    statuses\_count & 0.015544 &  \\
    \cline{1-2}
    favourites\_count & 0.011768 &  \\
    \cline{1-2}
    screen\_name\_length & 0.007641 &  \\
    \hline
    \end{tabular}}
  \label{tab:A2}%
\end{table}%

\begin{table*}[t]
\centering
\caption{Language that appears in MGTAB (ISO 639-1/639-2).}
\vspace{0.2cm}
\scalebox{0.9}{
\begin{tabular}{|l|l|l|l|l|l|l|l|l|l|}
\hline
ISO &Name     &ISO &Name      &ISO  &Name        &ISO &Name     &ISO &Name    \\
\hline
\hline
af	&AFRIKAANS &es &SPANISH	  &it	&ITALIAN	 &no &NORWEGIAN &sw	&SWAHILI   \\
\hline	
ar	&ARABIC	   &et &ESTONIAN  &ja	&JAPANESE    &pa &PUNJABI   &ta	&TAMIL     \\
\hline
bg	&BULGARIAN &fa &PERSIAN   &kn	&KANNADA	 &pl &POLISH    &te	&TELUGU    \\
\hline
bn	&BENGALI   &fi &FINNISH	  &ko	&KOREAN      &pt &PORTUGUESE&th	&THAI      \\
\hline
ca	&CATALAN   &fr &FRENCH    &lt	&LITHUANIAN  &ro &ROMANIAN  &tl	&TAGALOG   \\
\hline
cs	&CZECH	   &gu &GUJARATI  &lv	&LATVIAN     &ru &RUSSIAN   &tr	&TURKISH   \\
\hline
cy	&WELSH	   &he &HEBREW	  &mk	&MACEDONIAN  &sk &SLOVAK    &uk	&UKRAINIAN \\
\hline
da	&DANISH	   &hi &HINDI	  &ml	&MALAYALAM   &sl &SLOVENIAN &ur	&URDU      \\
\hline
de	&GERMAN	   &hr &CROATIAN  &mr	&MARATHI	 &so &SOMALI	&vi	&VIETNAMESE\\
\hline
el	&GREEK	   &hu &HUNGARIAN &ne	&NEPALI	     &sq &ALBANIAN  &zh	&CHINESE   \\
\hline
en	&ENGLISH   &id &INDONESIAN&nl	&DUTCH       &sv &SWEDISH	&   &          \\
\hline
\end{tabular}}
\label{tab:A4}
\end{table*}

\noindent
\textbf{Normalized of MGTAB.}
The MGTAB dataset contains the top 10 numerical and boolean features of each user's account, along with the tweet features extracted by LaBSE. The numerical features have undergone normalization using the MinMaxScaler method, which ensures that all features are on the same scale, preventing the model from being biased towards features with a larger range of values. The minimum and maximum values of each feature in MGTAB are available on the GitHub repository. This allows the trained model to be applied to new real-world data.

\noindent
\textbf{Licensing.}
The MGTAB dataset is governed by the CC BY-NC-ND 4.0 license. The implemented code within the evaluation framework is subject to the MIT license.

\noindent
\textbf{Potential negative societal impact.}
Although MGTAB datasets and evaluation frameworks are designed to facilitate stance detection and bot detection research, there is a risk that they could be misused by bot operators to analyze the characteristics of detection-evading bots and design more evasive bot algorithms.

\subsection{Feature Analysis}
\label{sec:FeatureAnalysis}
\noindent
\textbf{Information gain of features.}
Tab.~\ref{tab:A1} and Tab.~\ref{tab:A2} displays the top 10 information gain (IG) scores and corresponding boolean and numerical features used in user stance detection and bot detection, respectively.

\noindent
\textbf{Feature effectiveness analysis.}
The details of the user feature representations are shown in Tab.~\ref{tab:4}. Several studies have proposed various characteristics for account detection. To showcase the effectiveness of the extracted features in this paper, property features created from various literature~\cite{Alpher06,Alpher39,Alpher50} are utilized to compare the performance of different models based on the most frequently used friend and follower relationships~\cite{Alpher06}. In the experiment, we only use property features, and the results are shown in Tab.~\ref{tab:A3}.

\begin{table*}[htbp]
  \centering
  \caption{Details of user feature representations.}
    \vspace{0.2cm}
    \scalebox{0.9}{
    \begin{tabular}{|c|c|c|c|}

    \hline
    Features & Description & Type  & Dim \\
    \hline
    \hline
    profile\_use\_background\_image & If profile has background image & Boolean & 1 \\
    \hline
    default\_profile & If profile is set & Boolean & 2 \\
    \hline
    verified & If profile is verified & Boolean & 3 \\
    \hline
    followers\_count & Number of uers following this account & Numerical & 4 \\
    \hline
    default\_profile\_image & If profile image is default & Boolean & 5 \\
    \hline
    listed\_count & Public lists that use members of & Numerical & 6 \\
    \hline
    statuses\_count & Numbers of tweets and retweets & Numerical & 7 \\
    \hline
    friends\_count & Number of uers this account following & Numerical & 8 \\
    \hline
    geo\_enabled & Whether to enable geographical location & Boolean & 9 \\
    \hline
    favourites\_count & Number of this account likes & Numerical & 10 \\
    \hline
    created\_at & Time when the account was created & Numerical & 11 \\
    \hline
    screen\_name\_length & Length of screen\_name & Numerical & 12 \\
    \hline
    name\_length & Length of name & Numerical & 13 \\
    \hline
    description\_length & Length of description & Numerical & 14 \\
    \hline
    followers\_friends\_ratios & followers\_count/friends\_count & Numerical & 15 \\
    \hline
    default\_profile\_background\_color & If the profile background uses default color & Boolean & 16\\
    \hline
    default\_profile\_sidebar\_fill\_color & If the profile sidebar uses default color & Boolean & 17 \\
    \hline
    default\_profile\_sidebar\_border\_color & If the border of profile sidebar uses default color & Boolean & 18 \\
    \hline
    has\_URL & If URL is set & Boolean & 19 \\
    \hline
    profile\_background\_image\_URL & If the profile background image has URL & Boolean & 20 \\
    \hline
    tweet features & Averaged 768-dimensional features & Tweet & 21-788 \\
    \hline
    \end{tabular}}
  \label{tab:4}%
\end{table*}%

\begin{table*}[htbp]
\centering
\caption{The performance of using different features on MGTAB. Best results are highlighted in \textbf{bold}.}
\vspace{0.2cm}
\scalebox{0.9}{
\begin{tabular}{|c|l|c|c|c|c|c|c|c|c|}
\hline
\multirow{3}{*}{Task} & \multicolumn{1}{l|}{\multirow{3}{*}{Method}} & \multicolumn{8}{c|}{Features} \\
\cline{3-10}  &  & \multicolumn{2}{c|}{\cite{Alpher06}} & \multicolumn{2}{c|}{\cite{Alpher50}} & \multicolumn{2}{c|}{\cite{Alpher39}} & \multicolumn{2}{c|}{Ours} \\
\cline{3-10}  &  & Acc  & F1 & Acc  & F1 & Acc   & F1 & Acc   & F1 \\
\hline
\hline
\multirow{3}{*}{Stance}
& GCN & 63.8±0.1 & 63.0±0.4 & 61.1±0.5 & 60.0±0.5 & 62.7±0.7 & 61.5±1.2 & \textbf{65.2±0.5} & \textbf{64.9±0.4} \\
\cline{2-10}
& GAT & 70.2±0.2 & 69.1±0.5 & 70.3±0.1 & 69.5±0.3 & 70.7±0.5 & 69.9±0.5 & \textbf{70.9±0.6} & \textbf{70.1±0.5} \\
\cline{2-10}
& RGCN & 64.1±0.4 & 63.2±0.4 & 62.8±0.5 & 61.2±0.6 & 65.5±0.2 & 64.4±0.5 & \textbf{65.8±0.3} & \textbf{64.9±0.5} \\
\hline
\multirow{3}{*}{Bot}
& GCN & 71.1±0.5 & 65.8±0.8 & 71.7±0.4 & 66.8±0.6 & 78.0±0.6 & 68.9±0.5 & \textbf{79.5±0.2} & \textbf{71.1±0.1} \\
\cline{2-10}
& GAT & 68.1±0.2 & 61.0±0.3 & 68.5±0.4 & 63.7±0.6 & 73.0±0.4 & 69.8±0.4 & \textbf{73.7±0.3} & \textbf{71.5±0.4} \\
\cline{2-10}
& RGCN & 73.5±0.1 & 61.0±0.7 & 75.0±0.3 & 67.7±0.2 & 77.2±0.3 & 70.1±0.4 & \textbf{79.4±0.3} & \textbf{74.0±0.4} \\
\hline
\end{tabular}}
\label{tab:A3}
\end{table*}

\subsection{Impact of Different BERT Models}
\label{sec:BERTModels}
The 54 languages included in the MGTAB dataset are shown in Tab.~\ref{tab:A4}. In this section, we aim to demonstrate the effectiveness of encoding using LaBSE~\cite{Alpher54}. We adopt four pre-trained encoding models, LaBSE, RoBERTa~\cite{Alpher62}, SBERT~\cite{Alpher63}, and BART~\cite{Alpher64} to encode user tweets. The results using the above models to encode all tweets of users are shown in Tab.~\ref{tab:A5}. It can be observed that the detection performance of using LaBSE is better compared to other models. We infer that this is because noise is introduced when encoding multilingual text using the English pre-training model. LaBSE, on the other hand, can encode text in different languages into a shared embedding space, which is better suited to the multilingual text collected in this benchmark.

\subsection{Experiment Details}
\label{sec:ComputationDetails}
\noindent
\textbf{Experiment setup.}
In this paper, we utilize 2-layer GNNs and two fully connected layers for all GNN models. The input and output dimensions of the middle GNN layers are consistent, which are 64, 128, or 256. ReLU is used as the activation function, and the learning rate is set between 0.0001 to 0.01. The dropout rate is set from 0.3 to 0.5. For GAT, we set the number of attention heads as 8. For RGT, we set the number of transformer attention heads and semantic attention heads as 4. In S-HGN, $\beta$ is 0.05, and the rest remain at the default settings. We trained all GNN models for 300 epochs using Adam optimizer. For machine learning models, the number of estimators for AB and RF is set to 50 and 100, respectively. All experiments were conducted on a server with 9 TITAN RTX GPUs.

\noindent
\textbf{Datasets processing.}
For SemEval-2016 T6~\cite{Alpher21}, we extracted 20 largest features of IG: number of positive words, number of negative words, positive emotion counts, negative emotion counts, nouns words frequency, pronoun words frequency, verb words frequency, adjectives words frequency, number of special symbols, number of question mark, number of capital words, number of quoted words, retweet counts, mention counts, number of URLs, entropy of hastags, number of hashtags, and number of capitalized hashtags. For SemEval-2019 T7~\cite{Alpher25}, the feature was extracted by using RoBERTa~\cite{Alpher62}. For TwiBot-20~\cite{Alpher11}, we follow~\cite{Alpher06} for dataset processing and feature extraction. For Cresci-15~\cite{Alpher10}, Cresci-17~\cite{Alpher04}, and TwiBot-22~\cite{Alpher12}, we follow~\cite{Alpher12} for dataset processing and feature extraction.

\begin{table*}[t]
  \centering
  \caption{Performance using different encoding models on MGTAB. Best results are highlighted in \textbf{bold}.}
  \vspace{0.2cm}
  \scalebox{0.9}{
\begin{tabular}{|c|l|c|c|c|c|c|c|c|c|}
\hline
\multirow{3}{*}{Task} & \multirow{3}{*}{Method} & \multicolumn{8}{c|}{pretrain model} \\
\cline{3-10}      &       & \multicolumn{2}{c|}{RoBERTa} & \multicolumn{2}{c|}{SBERT} & \multicolumn{2}{c|}{BART} & \multicolumn{2}{c|}{Lasbe}\\
\cline{3-10}      &  & Acc & F1 & Acc & F1 & Acc & F1 & Acc & F1 \\
\hline
\hline
\multirow{6}{*}{Stance} & SVM  & 66.1±0.4 & 64.7±0.4 & 76.2±0.5 & 75.3±0.5 & 77.5±0.3 & 76.8±0.2 & \textbf{78.2±0.2} & \textbf{77.7±0.2} \\
\cline{2-10}  & DT  & 59.4±0.4 & 58.5±0.4 & 61.1±0.5 & 60.4±0.5 & 61.4±0.3 & 60.7±0.3 & \textbf{62.0±0.2} & \textbf{61.0±0.1}\\
\cline{2-10}  & RF  & 70.8±0.5 & 69.7±0.4 & 76.4±0.5 & 76.1±0.5 & 75.9±0.3 & 75.2±0.2 & \textbf{77.0±0.2} & \textbf{76.2±0.2}\\
\cline{2-10}  & GCN & 74.5±0.8 & 73.9±0.9 & 78.4±0.2 & 77.8±0.1 & \textbf{78.9±0.2} & \textbf{78.4±0.4} & 78.6±0.1 & 77.9±0.2 \\
\cline{2-10}  & GAT & 75.7±0.4 & 75.0±0.6 & 77.4±0.2 & 77.0±0.2 & 77.4±0.2 & 77.0±0.3 & \textbf{77.6±0.3} & \textbf{77.1±0.4} \\
\cline{2-10}  & BotRGCN & 76.4±0.4 & 75.5±0.5 & 77.9±0.3 & 77.5±0.2 & 79.0±0.1 & 78.6±0.2 & \textbf{79.2±0.1} & \textbf{78.7±0.2} \\
\hline
\multirow{6}{*}{Bot} & SVM  & 82.0±0.4 & 74.1±0.4 & 86.1±0.4 & 81.7±0.3 & 85.3±0.2 & 81.2±0.2 & \textbf{86.2±0.2} & \textbf{82.5±0.1} \\
\cline{2-10}   & DT  & 83.3±0.4 & 78.6±0.3 & 82.7±0.4 & 78.0±0.3 & 84.5±0.2 & 80.2±0.2  & \textbf{85.4±0.2} & \textbf{81.3±0.2}\\
\cline{2-10}   & RF  & 83.8±0.4 & 77.1±0.3 & 84.1±0.5 & 78.3±0.4 & 85.1±0.2 & 81.0±0.1  & \textbf{87.0±0.3} & \textbf{82.9±0.3} \\
\cline{2-10}   & GCN & 83.4±0.2 & 77.5±0.4 & 83.8±0.6 & 79.2±0.3 & 84.2±0.2 & 79.0±0.4 & \textbf{84.9±0.4} & \textbf{79.5±1.2} \\
\cline{2-10}   & GAT & 78.7±0.7 & 73.6±0.9 & 83.9±0.4 & 78.1±0.8 & 82.4±1.1 & 78.0±1.1 & \textbf{85.3±0.4} & \textbf{79.5±1.3} \\
\cline{2-10}   & BotRGCN & 84.3±0.4 & 79.3±0.7 & 86.1±0.1 & 81.5±0.1 & 85.7±0.1 & 80.9±0.3 & \textbf{87.2±0.1} & \textbf{83.2±0.3}\\
\hline
    \end{tabular}}
  \label{tab:A5}%
\end{table*}%

\end{document}